\documentclass[conference]{IEEEtran}
\IEEEoverridecommandlockouts
\usepackage{tabularray}

\usepackage{cite}
\usepackage{amsmath,amssymb,amsfonts}
\usepackage{algorithmic}
\usepackage{graphicx}
\usepackage{textcomp}
\usepackage{xcolor}
\usepackage{comment}
\usepackage{hyperref}
\usepackage{subcaption}
\usepackage{float}
\usepackage{tikz}
\usetikzlibrary{arrows.meta, positioning, shapes}
\usepackage[table]{xcolor}
\usepackage{multirow}
\usepackage[table]{xcolor}
\usepackage{multirow}
\usepackage[table,xcdraw]{xcolor}

\def\BibTeX{{\rm B\kern-.05em{\sc i\kern-.025em b}\kern-.08em
    T\kern-.1667em\lower.7ex\hbox{E}\kern-.125emX}}
\begin{document}

\title{Multitasking Embedding for Embryo Blastocyst Grading Prediction (MEmEBG) }

\author{
\IEEEauthorblockN{
Nahid Khoshk Angabini$^\star$, Mohsen Tajgardan$^\circ$, Mahesh Madhavan$^\star$, Zahra Asghari Varzaneh$^\star$\\
Reza Khoshkangini$^\star$, Thomas Ebner$^\dagger$
}

\IEEEauthorblockA{
$^\star$Department of Computer Science and Media Technology, Malmö University, Malmö, Sweden\\
$^\circ$Faculty of Electrical and Computer Engineering, Qom University of Technology, Iran\\
$^\dagger$Kepler Universitätsklinikum, Linz, Austria
}
}

\maketitle

\begin{abstract}
Reliable evaluation of blastocyst quality is critical for the success of in vitro fertilization (IVF) treatments. Current embryo grading practices primarily rely on visual assessment of morphological features, which introduces subjectivity, inter-embryologist variability, and challenges in standardizing quality assurance. In this study, we propose a multitask embedding-based approach for the automated analysis and prediction of key blastocyst components, including the trophectoderm (TE), inner cell mass (ICM), and blastocyst expansion (EXP). The method leverages biological and physical characteristics extracted from images of day-5 human embryos. A pretrained ResNet-18 architecture, enhanced with an embedding layer, is employed to learn discriminative representations from a limited dataset and to automatically identify TE and ICM regions along with their corresponding grades—structures that are visually similar and inherently difficult to distinguish. Experimental results demonstrate the promise of the multitask embedding approach and potential for robust and consistent blastocyst quality assessment.

\end{abstract}


\section{Introduction}
\label{sec:introduction}
In vitro fertilization (IVF) has become a cornerstone of modern reproductive medicine, offering an effective solution for infertility as well as a means to reduce the risk of hereditary disease transmission\cite{dyer2016international}. A central challenge in IVF is the early and reliable prediction of blastocyst quality, as timely assessment of key developmental characteristics directly influences embryo selection, transfer strategies, and clinical outcomes\cite{istanbul2011}. The IVF process consists of ovarian stimulation, oocyte retrieval, fertilization in a laboratory setting, and the subsequent culture and transfer of embryos\cite{elder2020}. Although a single IVF cycle typically lasts only a few weeks, it involves multiple tightly coordinated developmental stages, each of which can critically affect treatment success. Because many embryos fail to progress under in vitro conditions\cite{istanbul2011}, multiple embryos are cultured simultaneously, making early prediction of blastocyst development and quality essential for informed clinical decision-making\cite{meseguer2011}.

The introduction of time-lapse monitoring (TLM) incubators has significantly advanced embryo assessment by enabling continuous, non-invasive observation of early embryonic development\cite{cruz2012timing}. These systems acquire high-resolution images at fixed intervals while maintaining stable culture conditions, capturing detailed temporal information throughout embryo growth\cite{kirkegaard2012}. Most embryos reach the blastocyst stage approximately five days after fertilization, although developmental trajectories vary and many embryos arrest before this stage\cite{gardner2016}. Since only blastocysts are considered suitable for transfer or cryopreservation, the ability to predict both blastocyst formation and its structural quality components as early as possible is a key determinant of clinical success\cite{meseguer2011}.

TLM image sequences allow embryologists to evaluate embryos by tracking morphologic and morphokinetic changes over time \cite{wong2013time}. In clinical practice, blastocyst quality is primarily assessed through the appearance of its major structural components, including the trophectoderm (TE), the inner cell mass (ICM), and the degree of blastocyst expansion\cite{gardner2016}. These features are typically evaluated manually using images acquired around day five of development, when blastocyst morphology becomes clearly discernible\cite{istanbul2011}. However, such assessments are subjective, labor-intensive, and often performed late in development, limiting their utility for early decision-making and standardized embryo ranking\cite{racowsky2011}.

Recent advances in artificial intelligence, particularly deep learning, have demonstrated strong potential for automating and standardizing medical image analysis \cite{Zahra2025,mirzaeian2023}. In the context of IVF, AI-based methods have shown promise in improving the consistency and accuracy of embryo evaluation and selection, thereby reducing treatment cycles and patient burden\cite{vermilyea2020}. E.g, in \cite{soulaimani2025multimodal}, a multimodal deep learning approach was introduced that combines Vision Transformer–based image analysis with structured environmental data to predict embryo quality, enabling a more comprehensive and objective assessment of in-vitro fertilization outcomes. 
Nevertheless, many existing approaches focus solely on predicting overall blastocyst viability or rely exclusively on single images acquired at the blastocyst stage\cite{tran2019}, without explicitly modeling or analyzing the individual blastocyst components that define clinical grading, such as TE, ICM, and expansion\cite{kragh2021}. In addition to such limitations, practical constraints restrict the widespread use of existing AI systems. Continuous TLM imaging is not available in all IVF laboratories, and image sequences may be incomplete or sparsely sampled due to operational constraints\cite{kirkegaard2012}. This highlights the need for predictive frameworks that can effectively learn from limited image data collected across the standard five-day culture period, while still capturing meaningful developmental patterns relevant to blastocyst quality.

To address these challenges, we propose a multitask embedding-based learning framework for automated blastocyst quality assessment that leverages transfer learning as a shared visual backbone, embedding space, and multi-task head layers.  A pretrained deep neural network is used to extract compact and discriminative image embeddings from embryo images collected over the five-day culture period. These embeddings are then fed into a unified multitask prediction \cite{10.1007/978-981-99-8076-5_15,KHOSHKANGINI2023118716} head that simultaneously estimates TE quality, ICM quality, and blastocyst expansion. By jointly optimizing all prediction tasks within a shared embedding space, the model captures common developmental representations while preserving task-specific discriminative features. This design reduces training time and computational cost compared to training separate single-task models and enhances predictive accuracy and robustness by exploiting the intrinsic correlations among TE, ICM, and expansion. 

In this learning process, the knowledge of each task TE, ICM, and EXP can be transferred and applied to other tasks to improve the generalization performance. This enabled us to study extracting the transferred knowledge between the tasks. Furthermore, at the same time, the shared representation will yield shorter training time, allowing us to add more individual tasks by constructing only the specific layers and training only those layers.

The following main research question (RQ) further elaborates the investigative objectives of our proposed approach:

\begin{itemize}

\item \textbf{RQ1- Multi-task Embedding Learning:} To what extent could the task of embryo blastocyst grading be performed using transfer and multi-task embedding approach? 

\end{itemize}

Considering the research question, this study aims to address the aforementioned limitation by employing a multi-task embedding approach to enhance embryo blastocyst grading performance in IVF clinics. To answer RQ1, we leverage the shared representations extracted from a base ResNet-18 model and develop a multi-task predictive framework for the simultaneous estimation of multiple grading scores.


\section{Data Representation}
\label{sec_data}
The dataset introduced by Saeedi et al. \cite{saeedi2017automatic} consists of 249 human day-5 blastocyst images collected during standard IVF treatments and is designed to support quantitative analysis of blastocyst morphology at the component level. Each image contains a single complete blastocyst and may include realistic imaging artifacts such as granular cell fragments, floating particles, or partial views of nearby sibling embryos, reflecting real clinical conditions. All embryos were obtained from patients treated at the Pacific  Centre for Reproductive Medicine (PCRM) between 2012 and 2016, with approval from the Canadian Research Ethics Board.

\begin{figure}[H] 
\centering

\begin{subfigure}{0.48\columnwidth}  
    \centering
    \includegraphics[width=\linewidth, height=2.5cm, keepaspectratio]{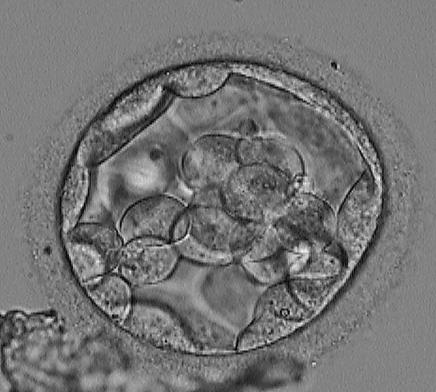}
    \caption{Original}
\end{subfigure}
\hfill
\begin{subfigure}{0.48\columnwidth}
    \centering
    \includegraphics[width=\linewidth, height=2.5cm, keepaspectratio]{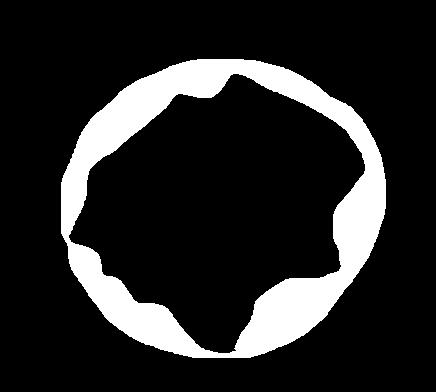}
    \caption{TE}
\end{subfigure}

\vspace{0.3cm}

\begin{subfigure}{0.48\columnwidth}
    \centering
    \includegraphics[width=\linewidth, height=2.5cm, keepaspectratio]{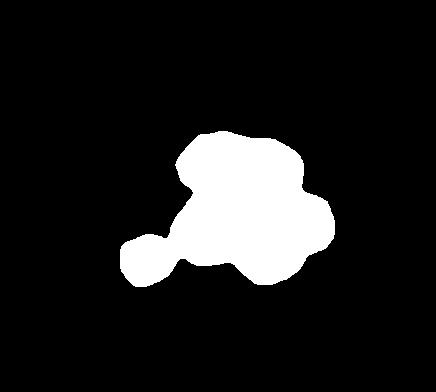}
    \caption{ICM}
\end{subfigure}
\hfill
\begin{subfigure}{0.48\columnwidth}
    \centering
    \includegraphics[width=\linewidth, height=2.5cm, keepaspectratio]{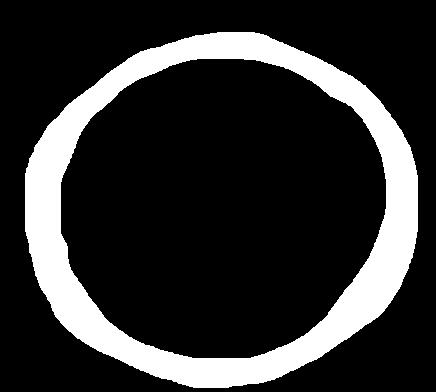}
    \caption{ZP}
\end{subfigure}

\caption{Example blastocyst image with segmented components: (a) original day-5 blastocyst, (b-d) corresponding masks highlighting the trophectoderm (TE), inner cell mass (ICM), and zona pellucida (ZP) regions.}
\label{fig:dataset_examples}
\end{figure}

For every blastocyst, expert embryologists manually annotated the inner cell mass (ICM) and trophectoderm (TE) regions, along with the zona pellucida (ZP) boundary. An example blasotcyst image with segmented components is shown in Figure \ref{fig:dataset_examples}. These annotations serve as ground truth for supervised learning and evaluation. In addition, each embryo is associated with a Gardner grading score, providing separate quality assessments for blastocyst expansion, ICM, and TE, as well as the clinical pregnancy outcome following embryo transfer. However, in this study, we focus on ICM, TE, and EXP components.

\section{Methodology}
\label{sec_approach}
\begin{figure*}[!h]
    \centering
    \includegraphics[width=0.80\textwidth]{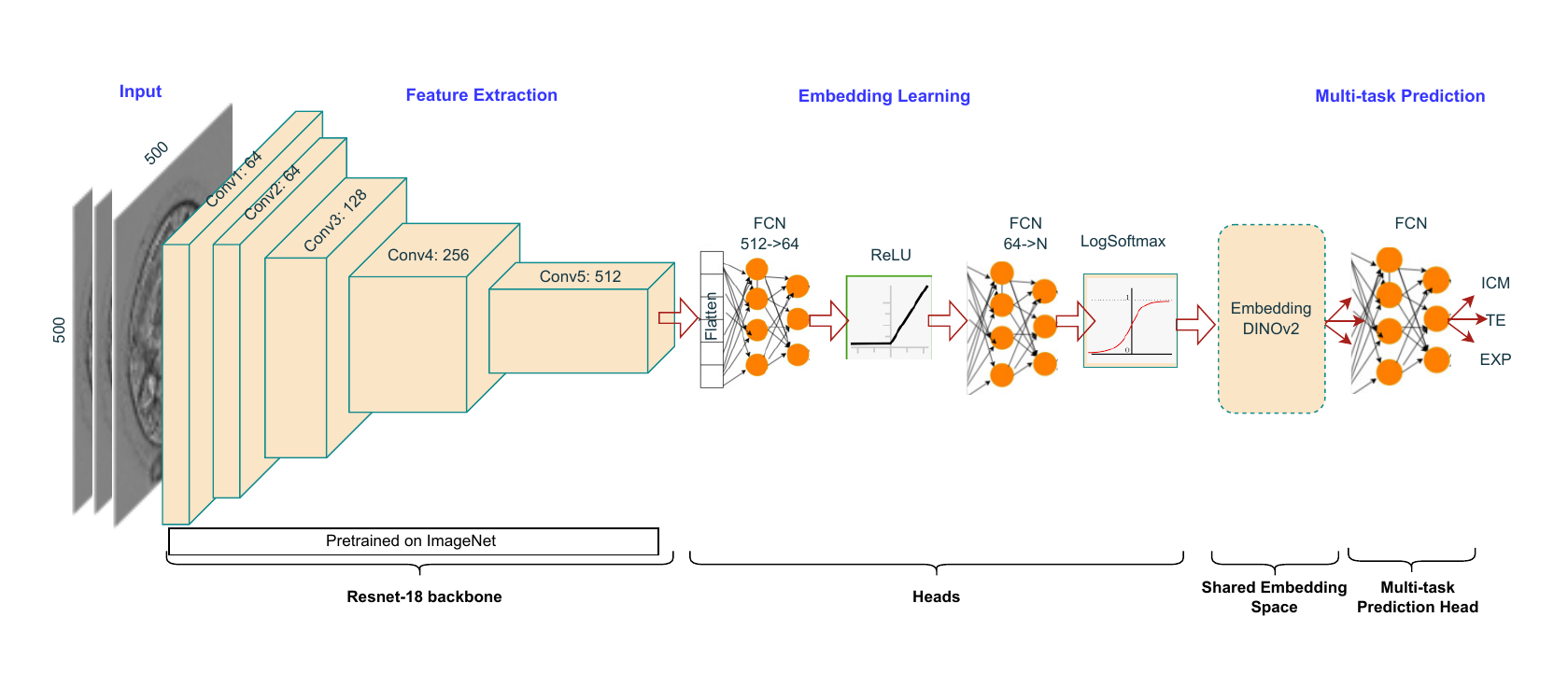}
    \caption{The conceptual view of the proposed multitask embedding approach} 
    \label{fig:proposed_approach} 
\end{figure*}

The conceptual overview of the proposed MEmEBG is illustrated in Figure \ref{fig:proposed_approach}. The approach is built upon a ResNet-18 base model with shared layers pretrained on ImageNet, followed by an embedding module exploiting DINOV2 and a set of fully connected layers with multiple task-specific heads. This architecture forms a multi-task network in which each head corresponds to a specific grading task.

\subsection{Multi-task Learning}

Multi-task learning (MTL) is a paradigm in which multiple related tasks are learned simultaneously within a single network, enabling the model to capture shared feature representations that enhance generalization and accelerate learning~\cite{zhang2021survey,10.1007/978-3-031-78107-0_15}. In our proposed system, individual tasks are denoted by $\tau_i$, and the complete set of tasks is $\mathcal{T} = {\tau_1, \tau_2, \dots, \tau_m}$.

Each task $\tau_i$ is associated with a loss function $L_{\tau_i}(\theta)$, representing the error for the corresponding embryo grading task, which is defined according to the blastocyst grading scores described in Section~\ref{sec_data}. MTL aims to jointly learn multiple grading criteria from a shared dataset $DS = {(x_j^i, y_j^i)}_{j=1}^{n_i}$, where $x_j^i$ denotes embryo image samples and $y_j^i$ their associated grading labels for task $\tau_i$.

The proposed model leverages shared parameters $\theta_{sh}$ to capture common morphological patterns across grading tasks, while task-specific parameters $\theta_i$ model individual grading characteristics. For each task, a task-dependent loss is computed as $L(\tau_i \mid \theta_{sh}, \theta_i)$.

\subsection{Feature Embedding and Multi-task Loss Function}
Features from all available modalities are first projected into a shared latent embedding space, parameterized by the shared parameters ($\theta_{sh}$). To obtain robust and informative visual representations, we employ DINOv2, a self-supervised Vision Transformer \cite{oquab2023dinov2}, to encode each embryo image into a high-dimensional embedding vector. The DINOv2 embedding space is particularly well-suited to this domain, as it captures fine-grained morphological structures, global developmental patterns, and semantic similarities between embryos without relying on manual annotations. This embedding constitutes the core representation of our multi-task architecture, providing a compact yet semantically rich description of embryo development from time-lapse images. 

On top of this shared embedding, a multi-task learning architecture is employed to address several embryo-related prediction objectives in parallel. Each task ($\tau_i \in \mathcal{T}$) corresponds to a distinct clinical or developmental outcome, here we refere to TE, ICM and EXP. For each task, a dedicated prediction head parameterized by ($\theta_i$) computes an individual loss ($L(\tau_i \mid \theta_{sh}, \theta_i)$), while all tasks share the same underlying embedding parameters ($\theta_{sh}$). This formulation encourages the shared embedding to capture task-invariant and biologically meaningful features, leading to improved generalization, increased data efficiency, and more coherent predictions across related embryo assessment tasks.

By jointly optimizing this shared embedding across all tasks in $(\mathcal{T})$, the model effectively exploits complementary information across modalities, while simultaneously attenuating modality-specific noise and task-dependent biases.

The overall learning objective is captured by a joint multi-task loss function over the combined parameters ${\theta_{sh}} \cup {\theta_i \mid \tau_i \in \mathcal{T}}$, expressed as:

\begin{equation}\label{eq:mtl_function}
L_{all}(X, \theta_{sh}, \theta_{i=1,\dots,m})= \sum_{i=1}^{m} L(\tau_i; \theta_{sh}, \theta_i)
\end{equation}

This joint optimization promotes knowledge transfer among related tasks, improves generalization in data-limited IVF settings, and produces biologically consistent, clinically meaningful embryo representations.

\section{Experimental Results}
\label{sec:result}
\setlength{\tabcolsep}{0.9pt}

\begin{table}[t]
\small
\centering

\caption{Performance comparison between single-task learning (STL) and multitask learning (MTL) models for score grading prediction. Statistical significance was evaluated using the $5\times2$ cross-validated paired t-test.}

\begin{tabular}{|l|cc|cc|cc|}
\hline
 & \multicolumn{2}{c|}{\textbf{TE}} & \multicolumn{2}{c|}{\textbf{ICM}} & \multicolumn{2}{c|}{\textbf{EXP}} \\ \cline{2-7}

\textbf{Method} &
\begin{tabular}[c]{@{}c@{}}f-score\\ \& std\end{tabular} & p-value &
\begin{tabular}[c]{@{}c@{}}f-score\\ \& std\end{tabular} & p-value &
\begin{tabular}[c]{@{}c@{}}f-score\\ \& std\end{tabular} & p-value \\ \hline

\textbf{STL-TE}  & $0.60 \pm 0.03$ & $<0.05$ & --- & --- & --- & --- \\ \hline
\textbf{STL-ICM} & --- & --- & $0.64 \pm 0.03$ & 0.1 & --- & --- \\ \hline
\textbf{STL-EXP} & --- & --- & --- & --- & $0.72 \pm 0.04$ & 0.05 \\ \hline

\rowcolor{gray!15}
\textbf{MTL-TE}  & $0.64 \pm 0.02$ & --- & --- & --- & --- & --- \\ \hline

\rowcolor{gray!15}
\textbf{MTL-ICM} & --- & --- & $0.63 \pm 0.12$ & --- & --- & --- \\ \hline

\rowcolor{gray!15}
\textbf{MTL-EXP} & --- & --- & --- & --- & $0.76 \pm 0.02$ & --- \\ \hline

\end{tabular}
\label{tbl:comparison}
\end{table}

The results and evaluations are presented in relation to the research question introduced in Section~\ref{sec:introduction}. RQ1: To what extent can embryo blastocyst grading be performed using transfer learning and a multi-task embedding approach?

To address RQ1, we conducted a series of experiments focused on embryo grading prediction. Approximately 75\% (211 samples) of the dataset was used for training a fully connected multi-head network, while the remaining 25\%  (50 samples) was reserved for testing. First, we developed multiple predictive models at the single-task learning (STL) level, where each model was trained independently to predict an individual grading score. This allowed us to evaluate the performance of standalone models for each grading task.

Next, we designed a multi-task learning (MTL) embedding network (see Figure \ref{fig:proposed_approach}) that jointly trained all grading tasks within a single learning process. The proposed network leveraged a shared embedding space while preserving task-specific parameters, as formalized in Equation~\ref{eq:mtl_function}. Both the STL and MTL models employed ResNet19 as the backbone architecture.

These experiments served two main purposes. First, they evaluated whether framing embryo grading as a multi-task learning problem leads to improved predictive performance compared to single-task models. Second, they established baseline results to benchmark the effectiveness of the proposed MTL approach.

Table \ref{tbl:comparision} summarizes the performance of the implemented predictive models for automated embryo gradin including TE, ICM, and EXP. Due to the imbalance of the data and classes in this study, we focused on f-score metric to assess the predictive models. As shown, the multitask learning models consistently outperform the corresponding single-task models for both TE and EXP grading. Specifically, multitask models achieve an average performance of 0.64 versus 0.60 for TE prediction and 0.76 versus 0.72 for EXP prediction, demonstrating a clear advantage in jointly learning related embryological features.

Statistical hypothesis testing conducted at a significance level of 
$\alpha =0.05$ confirms that the observed performance improvements for TE and EXP are statistically significant, indicating that the gains achieved by the multitask approach are unlikely to be due to random variation across cross-validation folds. These results suggest that shared representations across embryo grading tasks enable the model to better capture correlated morphological and developmental patterns, particularly those influencing blastocyst expansion and trophectoderm quality.

In contrast, for ICM grading, the single-task model exhibits slightly higher predictive performance compared to the multitask model. However, this improvement does not reach statistical significance, as the null hypothesis could not be rejected. Furthermore, performance variability across cross-validation folds indicates that the observed difference is not consistent or robust. This finding suggests that ICM grading may rely on more task-specific visual cues, which are less effectively leveraged through shared multitask representations under the current model configuration.

Overall, these results highlight the benefits of multitask learning for embryo grading prediction, particularly for TE and EXP, while also underscoring the need for task-aware architectural or loss-weighting strategies when modeling ICM to ensure balanced learning across all embryological grading criteria.

\begin{figure}
    \centering
    \includegraphics[width=0.3\textwidth]{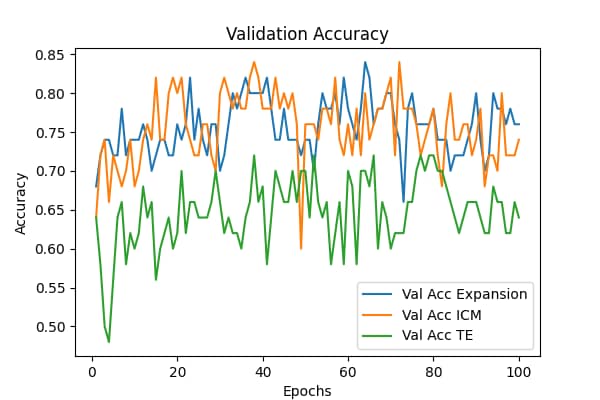}
    \caption{The validation accuracy over the training process for ICM, TE, and EXP. } 
    \label{fig:val_accuracy} 
    
\end{figure}

\begin{figure*}[htbp]
    \centering
    \includegraphics[width=0.7\textwidth]{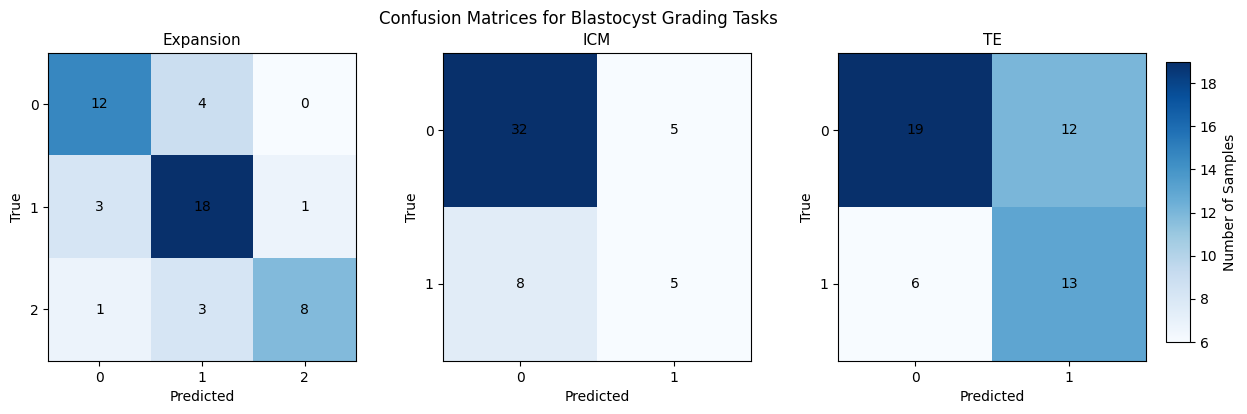}
    \caption{The confusion matrix from three different tasks.} 
    \label{fig:val} 
    
\end{figure*}

To better understand the behavior of the multitask embedding approach with respect to individual embryo grading outcomes, we further analyzed the per-grade prediction performance for each grading category. This fine-grained analysis allows us to assess how effectively the shared embedding captures class-specific characteristics across different embryo quality levels.

As illustrated in Table \ref{tab:icm_report}, which reports results for ICM Grades A and B, the multitask model demonstrates strong performance for Grade A, achieving an F-score of 0.85, while performing considerably worse for Grade B, with an F-score of 0.44. A similar performance pattern is observed for trophectoderm (TE) grading, as shown in Table \ref{tab:te_report}, where higher-grade embryos are predicted more reliably than lower-grade counterparts.

This imbalance in predictive performance is likely attributable to the limited number of samples per grade in the training dataset, particularly for underrepresented categories such as Grade B. The same constraint is also present in the test dataset, further amplifying the difficulty of learning robust decision boundaries for minority classes. As a result, the multitask embedding appears to favor dominant grading patterns, leading to reduced generalization performance for less frequent grades.

\begin{table}[htbp]
\centering
\caption{Expansion Classification Performance}
\label{tab:expansion_report}
\begin{tabular}{ccccc}
\hline
\textbf{Class} & \textbf{Precision} & \textbf{Recall} & \textbf{F1-score} & \textbf{Support} \\
\hline
0 & 0.75 & 0.73 & 0.75 & 16 \\
1 & 0.80 & 0.87 & 0.76 & 32 \\
2 & 0.89 & 0.67 & 0.76 & 12 \\
\hline
Accuracy &  &  & 0.76 & 50 \\
Macro Avg & 0.78 & 0.76 & 0.76 & 50 \\
Weighted Avg & 0.78 & 0.76 & 0.76 & 50 \\
\hline
\end{tabular}
\end{table}

\begin{table}[htbp]
\centering
\caption{ICM Classification Performance.}
\label{tab:icm_report}
\begin{tabular}{ccccc}
\hline
\textbf{Class} & \textbf{Precision} & \textbf{Recall} & \textbf{F1-score} & \textbf{Support} \\
\hline
A & 0.80 & 0.84 & 0.85 & 37 \\
B & 0.50 & 0.38 & 0.44 & 13 \\
\hline
Accuracy &  &  & 0.76 & 50 \\
Macro Avg & 0.65 & 0.61 & 0.64 & 50 \\
Weighted Avg & 0.72 & 0.74 & 0.73 & 50 \\
\hline
\end{tabular}
\end{table}

\begin{table}[htbp]
\centering
\caption{TE Classification Performance.}
\label{tab:te_report}
\begin{tabular}{ccccc}
\hline
\textbf{Class} & \textbf{Precision} & \textbf{Recall} & \textbf{F1-score} & \textbf{Support} \\
\hline
A & 0.76 & 0.61 & 0.68 & 31 \\
B & 0.52 & 0.68 & 0.60 & 19 \\
\hline
Accuracy &  &  & 0.66 & 50 \\
Macro Avg & 0.64 & 0.65 & 0.64 & 50 \\
Weighted Avg & 0.67 & 0.66 & 0.65 & 50 \\
\hline
\end{tabular}
\end{table}


Additional insights into the learning dynamics are provided by the training and validation accuracy trends shown in Figure \ref{fig:val_accuracy}. During the first 100 training epochs, the validation accuracy exhibits noticeable variability, suggesting that the multitask setting introduces a more challenging optimization landscape. This behavior is particularly evident in the TE prediction task, where fluctuations are observed and accuracy occasionally falls below 70\%. Rather than indicating a fundamental shortcoming, these oscillations highlight the sensitivity of multitask learning to data availability and task imbalance, especially in the presence of limited training samples.

Overall, the results indicate that the multitask embedding approach effectively captures shared representations for well-represented embryo grades and provides a meaningful inductive bias across related grading tasks. Even when the F-score remains modest, the model demonstrates the ability to exploit cross-task dependencies that would be difficult to leverage in single-task settings. The observed limitations are primarily attributable to class imbalance and restricted sample sizes in lower-frequency grading categories, rather than inherent weaknesses of the multitask framework. Importantly, incorporating segmentation techniques alongside the embedding representation—by explicitly focusing the model on biologically relevant regions of the embryo—has the potential to further support feature learning and reduce noise in the shared representation space. Such spatially informed embeddings are expected to improve task-specific discrimination and stability during training, ultimately leading to enhanced performance and more robust convergence in embryo grading prediction.

\section{Conclusion}
In conclusion, this study demonstrates that a multitask embedding-based approach can effectively analyze key blastocyst components, including the TE, ICM, and EXP. Despite the limited size of the training dataset—approximately 150 images—the method successfully captures meaningful shared representations across tasks, demonstrating the robustness of multitask learning even under challenging data-scarce conditions. By leveraging the biological and physical characteristics of day-5 human embryos, the approach enables automatic, objective, and quantitative identification of TE and ICM regions, which are otherwise visually difficult to distinguish.

The results underscore the potential of this system to enhance embryo assessment, particularly when integrated with pregnancy outcome data. Furthermore, incorporating segmentation techniques alongside the embedding representation offers a promising avenue to further improve feature learning, task-specific performance, and training stability. Overall, these findings highlight the promise of multitask learning as a powerful tool for advancing reproductive technologies, providing a foundation for more accurate and robust embryo evaluation in future clinical applications.

\section*{Acknowledgment}
\footnotesize This study is conducted as part of the EIVF-AI project funded by Vinnova, the
Swedish Governmental Agency for Innovation Systems (Grant No.2024-01462).

\bibliographystyle{ieeetr}  
\bibliography{references}

\end{document}